# VITAL: Vision Transformer Neural Networks for Accurate Smartphone Heterogeneity Resilient Indoor Localization

Danish Gufran, Saideep Tiku, Sudeep Pasricha
*Department of Electrical and Computer Engineering*
*Colorado State University*
Fort Collins, CO, United States
{danish.gufran, saideep, sudeep}@colostate.edu

*Abstract* – Wi-Fi fingerprinting-based indoor localization is an emerging embedded application domain that leverages existing Wi-Fi access points (APs) in buildings to localize users with smartphones. Unfortunately, the heterogeneity of wireless transceivers across diverse smartphones carried by users has been shown to reduce the accuracy and reliability of localization algorithms. In this paper, we propose a novel framework based on vision transformer neural networks called VITAL that addresses this important challenge. Experiments indicate that VITAL can reduce the uncertainty created by smartphone heterogeneity while improving localization accuracy from 41% to 68% over the best-known prior works. We also demonstrate the generalizability of our approach and propose a data augmentation technique that can be integrated into most deep learning-based localization frameworks to improve accuracy.

## I. INTRODUCTION

Indoor localization systems (ILS) allow the localization of items or persons within buildings and subterranean facilities (e.g., underground tunnels, mines). As the Global Positioning System (GPS) is unreliable in indoor settings due to the lack of visible contact with GPS satellites, an ILS must rely on alternative means of localization. These include techniques such as triangulation and fingerprinting with wireless technologies, e.g., Wi-Fi, Bluetooth Low Energy (BLE), Ultra-Wideband (UWB), and Radio Frequency Identification (RFID) [1]. Such ILS techniques are now powering several geo-location-based embedded systems and business platforms including Amazon warehouses robots, parking features in self-driving cars, IoT indoor navigation solutions, and so on [2].

Out of the available approaches, fingerprinting has shown to be one of the most scalable, low-cost, and accurate indoor localization solutions [3], [4]. The operation of fingerprint-based ILS is divided into two phases. The supplier of the ILS gathers received signal strength indication (RSSI) data for wireless Access Points (APs), such as Wi-Fi APs, at various indoor locations of interest during the first (offline) phase. The unique RSSI values at each location form a "fingerprint" vector at that location, which is different from the fingerprint vector at other locations. Subsequently, a database of RSSI fingerprint vectors and their corresponding locations is created. This database can be used together with pattern matching algorithms to estimate location in the second (online) phase, where a person uninformed of their location captures an RSSI fingerprint using their smartphone and sends it to the algorithm (on the smartphone or a cloud server) to predict their location. This predicted location can be displayed on the smartphone for real-time location visualization.

In recent years, Wi-Fi-based fingerprinting when combined with machine learning has been shown to provide promising accuracy for indoor location predictions [5]. Neural networks can capture prominent features in RSSI values from Wi-Fi signals captured by a smartphone and map these features to the location of the smartphone, which is often referred to as the Reference Point (RP) prediction problem. Despite the claimed benefits of Wi-Fi fingerprinting-based indoor localization, there remain several unresolved issues. Wi-Fi signals are affected by poor wall penetration, multipath fading, and shadowing. Due to these difficulties, establishing a deterministic mathematical relationship between the RSSI and distance from APs is challenging, especially in dynamic environments and across different device configurations. The latter problem is particularly challenging as different smartphones use different wireless transceivers, which significantly changes RSSI values captured by the different smartphones at the same location.

The performance of indoor localization solutions today is very susceptible to heterogeneity within smartphones, as well as other embedded and IoT devices that may participate in ILS. We quantify the impact of this device heterogeneity on several smartphone devices in Section III. This device heterogeneity leads to unpredictable variation in performance (location estimates) across users that may be present at the same physical location. The variations reduce precision, and ultimately the accuracy of ILS.

In this paper, we present VITAL, a novel embedded software framework that enables robust and calibration-free Wi-Fi-based fingerprinting for smartphones, with broad applicability to embedded and IoT devices used in ILS. VITAL aims to achieve device invariance with the end goal of achieving negligible precision error across heterogeneous smartphones and high-accuracy localization. Our framework employs vision transformer neural networks for the first time to the problem of Wi-Fi fingerprinting-based indoor localization. RSSI data is converted to images and processed using novel algorithms that integrate the transformer neural network to assist with device heterogeneity-resilient indoor localization.

The main contributions of our work are:

- We present a novel vision transformer neural network-based indoor localization solution that is resilient to device heterogeneity and achieves high-accuracy localization,
- We propose a novel RSSI image-based model to capture prominent features from Wi-Fi RSSI fingerprints,
- We design a data augmentation technique that can be seamlessly integrated into any deep learning model, including vision transformers, to improve heterogeneity resilience,
- We evaluate the localization performance of VITAL against the best-known indoor localization solutions from prior work, across several real indoor paths in multiple buildings, and for various smartphones carried by users.

## II. RELATED WORKS

Indoor localization competitions in recent years held by Microsoft [6] and NIST [7] have advocated for Wi-Fi fingerprinting-based indoor localization as the foundation for future ILS. The improvements in the computational abilities of smartphones and embedded platforms have empowered them to execute deeper and more powerful machine learning (ML) models to improve indoor localization performance. Thus, many efforts in recent years have started to explore the use of ML with Wi-Fi fingerprinting.

A survey of ML-based Wi-Fi RSSI fingerprinting schemes, including data preprocessing and ML prediction models for indoor localization, was presented in [8]. In [9], an analysis was performed to determine how to select the best RSSI fingerprints for ML-based indoor localization. In [10] and [11], Deep feedforward Neural Networks (DNNs) and Convolutional Neural Networks (CNNs) were shown to improve indoor localization accuracies over classical localization algorithms. A recent work in [12] explored the traditional transformer neural network on Channel State Information (CSI) data

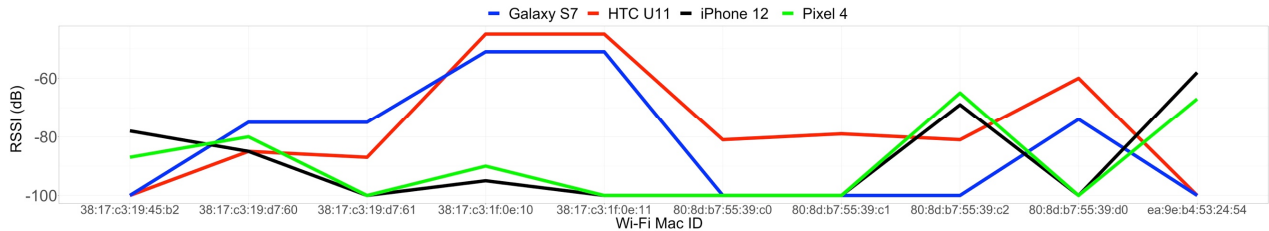
Figure 1: RSSI values of ten Wi-Fi APs observed by four different smartphones at the same location. The values differ significantly across devices.

for indoor localization. The work in [13] proposes an early-exit deep neural network strategy for fast indoor localization on mobile devices. These prior works motivate the use of ML-based techniques for indoor localization. However, none of these efforts address real-world challenges related to device heterogeneity.

The work in [14] performed a comparative study on the behavior of classification and regression-based ML models and their ability to deal with device heterogeneity. In [15], an unsupervised learning mechanism with Generative Adversarial Networks (GAN) was used for data augmentation with deep neural networks. However, this work does not address localization with noisy fingerprint data (e.g., fluctuating RSSI data from different smartphones). The work in [16] employed a deep autoencoder for dealing with noisy fingerprint data and device heterogeneity. The deep autoencoder was used along with traditional K-Nearest-Neighbor (KNN) classification models, but results for the approach across smartphones are not promising. Calibration-free approaches were proposed that used Signal Strength Difference (SSD) and Hyperbolic Location Fingerprints (HLF) with pairwise ratios to address heterogeneity [18]. However, both these approaches also suffer from slow convergence and high errors across larger sets of diverse smartphones.

The recent works ANVIL [19], SHERPA [20], CNNLoc [21], and WiDeep [22] improve upon the efforts mentioned above, to more effectively cope with device heterogeneity while maintaining lower localization errors. In, [19] a multi-headed attention neural network-based framework was proposed for indoor localization. The work in [20] combined KNN and deep neural networks for fingerprint classification. In [21], CNNs were used for regression-based localization prediction. The work in [22] employed a stacked autoencoder for data augmentation and a Gaussian process classifier for localization classification. The ML models and augmentation techniques in all of these works aim to improve resilience to device heterogeneity while minimizing localization error.

As [19]-[23] represent the state-of-the-art heterogeneity-resilient indoor localization efforts, we modeled these frameworks and used them to contrast against our framework. Deep learning models used for indoor localization are typically very large in size, consisting of millions or even billions of parameters, which can make them challenging to deploy on resource-constrained devices [24]. Model compression techniques can be used to reduce the size of these models, making them more efficient to deploy in real-world applications [25]. This is particularly important for indoor localization, where accurate and efficient models are necessary to locate users within indoor environments. The works in [24], [25] demonstrate strategic approaches in reducing the model dimensionality without compromising the model's performance. Mobile devices are also vulnerable to security threats such as data breaches and unauthorized access [26]. Overcoming security vulnerabilities in deep learning-based indoor localization frameworks on mobile devices is crucial to ensure the privacy and security of user data [26].

Our proposed VITAL framework integrates a novel data augmentation approach with an enhanced vision transformer neural network to achieve promising results for resilience towards device heterogeneity, while maintaining extremely low localization errors, as shown in Section VI.

III. ANALYSIS OF RSSI FINGERPRINTS

To illustrate the challenge due to device heterogeneity during Wi-Fi fingerprinting-based indoor localization, we present RSSI values captured by four different smartphones at a single location. These devices are a subset of those used to evaluate our proposed framework against the state-of-the-art, with results in Section VI. Figure 1 shows a plot of the RSSI from the four smartphones. The RSSI values captured are in the range of –100dB to 0dB, where –100dB indicates no visibility and 0dB is the strongest signal. The figure shows 10 RSSI values from 10 different Wi-Fi APs captured at a single location. The solid lines represent mean RSSI values across 10 samples taken for each smartphone at that location. Note that the RSSI values are combined into a "fingerprint" vector for that location. From the figure, we can make the following observations:

- The RSSI values captured by the different smartphones show deviations from each other, which is referred to as the AP visibility variation problem in indoor localization. Our proposed framework is curated to address this problem.
- There are similarities in RSSI patterns between some device pairs, such as between HTC-U11 and Galaxy-S7, and between iPhone-12 and Pixel-4. Even though there is variation across the patterns, it is possible to develop a calibration-free deep learning model, to learn to predict locations for these patterns.
- The skews of RSSI variations, even among the pairs of devices with similar patterns, are not fixed. This complicates the learning problem. To overcome this, we propose a powerful image-based fingerprinting technique that converts the RSSI fingerprints to an image that emphasizes high-visibility APs.
- Some APs that are visible to a smartphone may not be visible with a different smartphone at the same location. For example, the WiFi AP with a MAC ID (80:8d:b7:55:39:c1) is only visible to the HTC-U11 device (–81dB value) but is not visible for the other smartphones (–100 dB value). This is referred to as the missing APs problem in indoor localization. We develop a novel data augmentation module to address this challenge.

We analyzed RSSI patterns across multiple locations in different buildings and with additional smartphones (results for which are presented in Section VI), and our observations were consistent with the discussion above. The observed discrepancies in RSSI readings motivated our design of the novel VITAL framework that improves upon the state-of-the-art in fingerprinting-based indoor localization.

IV. VISION TRANSFORMERS: OVERVIEW

A transformer is a powerful deep neural network model that relies on the self-attention mechanism [17]. The self-attention mechanism mimics human cognitive attention, enhancing some parts of the input data while diminishing other parts, via a differential weighting approach. The motivation for doing so is to force the neural network to devote more focus to the small, but important, parts of the data. Learning which part of the data is more important than another depends on the context, which is trained by gradient descent. Transformers are today being widely used in Natural Language Processing (NLP) tasks.

Since the original transformer was proposed, many efforts have attempted to go beyond NLP and apply it to computer vision tasks. Unfortunately, the naive application of self-attention to images requires that each pixel attends to every other pixel. With quadratic cost in the number of pixels, this does not scale to realistic input sizes. The vision transformer (ViT) model [27] was recently proposed as a scalable approach to applying transformers to image processing. ViT improves upon the traditional transformer model which lacks some of the image-specific inductive biases (such as translation equivariance and locality) that make convolutional neural networks (CNNs) very effective for computer vision tasks.

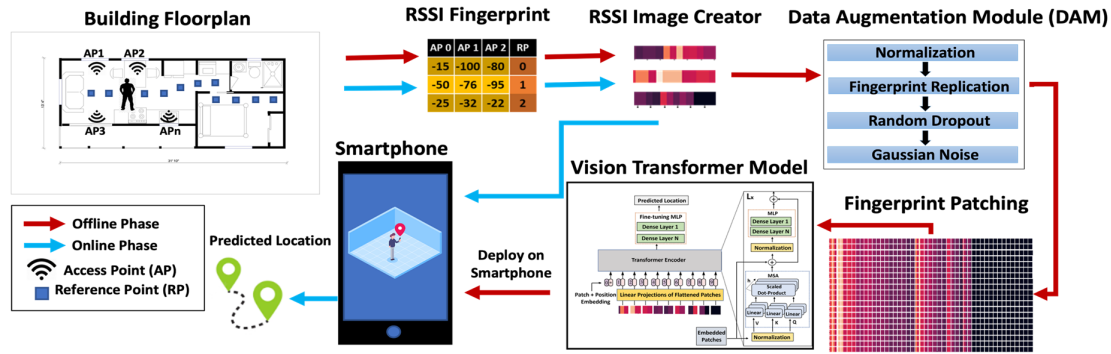

Figure 3: An overview of the proposed device heterogeneity-resilient VITAL indoor localization framework.

An overview of the ViT model is depicted in Figure 2. ViT consists of three main components: 1) position embedding of patches, 2) transformer encoder block, and 3) a Multi-Layer Perceptron (MLP) head. The first component of ViT creates patches from an image, essentially dividing the input image into small chunks and then positionally encoding each chunk to retain the order of the patches created. These positionally encoded patches are referred to as embedded patches and allow the ViT to be permutationally invariant. The embedded patches are subsequently sent as inputs to the transformer encoder, which is the second component. The layers of the transformer encoder are shown in Figure 2 and consist of $L$ alternating layers of a multi-headed self-attention (MSA) and an MLP block, in addition to layer normalizations, and residual connections. In the MSA block $h$ denotes the number of heads. The output of the transformer is sent to the third component, which is an MLP head that contains two layers with a GELU non-linearity for classification.

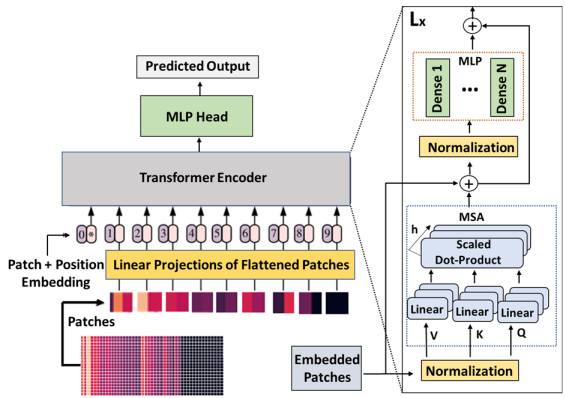

Figure 2: Overview of Vision Transformer Neural Network

## V. VITAL FRAMEWORK

Figure 3 shows an overview of our proposed VITAL framework that adapts (and enhances) vision transformers for the indoor localization problem. The proposed framework consists of 2 phases, namely the offline phase, depicted with red arrows, and the online phase, depicted with blue arrows. In the offline phase, the RSSI fingerprints are captured at each RP, with different smartphones, across various buildings. We use three fingerprint values at each RP and create a 1D image with three channels using a custom RSSI image creator. This module maps the three RSSI readings for each AP to a pixel. Thus, a pixel represents the three RSSI values for an AP, and the fingerprint vector at an RP consists of RSSI readings for all APs, in the form of a 1D image. The 1D image is pre-processed with a data augmentation module (DAM) and represented as 2D images. These images are then sent as inputs to a vision transformer model, to learn the mapping between the fingerprints and RP locations. In the online phase, the user requesting their location captures RSSI fingerprints from the smartphone at an unknown location. The RSSI fingerprints are pre-processed with DAM and represented as an image, as done in the offline phase. The image is then sent to the trained vision transformer to predict the location.

In the following subsections, we describe the two main components of our framework: DAM and the vision transformer.

### A. Data augmentation module

Our data augmentation module (DAM) is responsible for preparing the input data to help improve the sensitivity of location prediction and make it calibration-free. DAM consists of four stages as shown in Figure 3: data normalization, fingerprint replication, random dropout, and gaussian noise.

The first stage normalizes the input data. This is an important step as it ensures that each pixel has a similar distribution, making the convergence faster during training. The layer standardizes the features in the fingerprint to achieve smoother gradients and better generalization during training. The second stage replicates the fingerprint data to concatenate the augmented features with the original features as a single image. The replication is done such that the input 1D image size of $1 \times R$ (where $R$ is the number of RPs) is augmented and represented as a 2D image of size $R \times R$. We explored various image augmentation sizes to determine the optimal value, and this study is presented in Section VI.B. The third and fourth steps involve the addition of random dropout and Gaussian noise. The random dropout is used to randomly drop some APs or features in the input data to represent missing APs and allow the model to be robust to missing APs when making location predictions. The Gaussian noise is used to infill the dropped features with some random noise to represent different AP visibilities. This allows capturing the effect of fluctuating RSSI values due to dynamic environmental factors in indoor locations and makes the trained model more robust to RSSI variations.

In our framework, DAM is applied after the fingerprint capture step. This module can be integrated into any ML framework to improve robustness of indoor localization. In fact, we demonstrate in Section VI.D that DAM can have a notable positive impact for many indoor localization frameworks from prior work.

### B. Vision transformer

The output from DAM is sent to a vision transformer model that we modify and adapt for the indoor localization problem domain. We modified some of the layers from the vision transformer model proposed in [27], and make changes to the image creation process, input parameters for the MSA, and the MLP layers in the transformer encoder and the classification head, as discussed below.

Post augmentation from DAM, we now have a 2D image of size $R \times R$. This image is sliced into small patches of size $P \times P$. We explored the impact of different patch sizes, with results presented in Section VI.B. The number of patches per image is represented as $N$, where $N$ is calculated by dividing the area of the input image ($H*W$) by the area of the patch sizes, i.e., $N = (H*W)/(P*P)$. The $N$ patches serve as an input sequence to the vision transformer. As the model expects a constant vector of size $N$ for all its layers, to satisfy this requirement, we flatten the patches and map them as a linear trainable projection. These projections are then sent to the transformer encoder block.

The self-attention sub-block in the transformer encoder calculates a weighted average for each feature in the projections from the input image. The weights are proportional to the similarity score between each feature. The attention mechanism in the sub-block is modeled such that it extracts values ($V$) with associated keys ($K$) for a given query ($Q$). Through this, we calculate attention as:

$$Attention(Q,K,V) = softmax(Dot\ Score)V \quad (1)$$

$$Dot\ Score = \left(Q\ K^T / \sqrt{d_k}\right) \quad (2)$$

$$Q = XW^Q, K = XW^K, V = XW^V \quad (3)$$

where $d_k$ is the dimensionality of the key, $X$ is the input, and $W^Q$, $W^K$, and $W^V$ are the trainable weight matrices for $Q$, $K$, and $V$, respectively. In our framework, $Q$ is the patched images, $K$ is the one-hot encoded positions of each patch, and $V$ is the one-hot encoded RP locations of the input image. The higher the dot product score, the higher the attention weights will be, which is why it is considered a similarity measure. Using this information, we implement a multi-headed self-attention (MSA) mechanism to extract information from different input representations. Multi-head attention allows the model to jointly attend to information from different representation subspaces at different positions. With a single attention head, averaging inhibits this. The multi-headed attention function is expressed as:

$$MultiHead(Q,K,V) = Concat(h_1, h_2, ... h_n)W^o$$
$$where\ h_i = Attention(QW_i^Q, KW_i^K, VW_i^V) \quad (4)$$

We use sensitivity analysis to determine the optimal number of MSA heads in Section VI.B. The encoder comprises an MSA sub-block and an MLP sub-block with two dense neural network layers and a GELU non-linearity. We used layer normalization before each MSA and MLP sub-block to independently normalize inputs across all features, and concatenated the MSA sub-block output with the MLP sub-block outputs to restore any lost features. The encoder outputs a vector of values that corresponds to the location class of the input RSSI image. To make the indoor localization framework calibration-free, we fine-tuned the output from the transformer encoder using a fine-tuning MLP block with a series of MLP layers. We explored the optimal number of dense layers via sensitivity analysis (Section VI.B), and set the last dense layer to contain as many neurons as the number of RPs.

After training, we deployed the model on smartphones for online inference. To predict the unknown location from RSSI values collected at an unknown location, we pre-processed and represented the RSSI fingerprints as images, as done in the offline phase, and sent the requesting image to our vision transformer neural network.

The VITAL framework is set to be calibration-free by group training the neural network model. The group training combines RSSI fingerprint data from different smartphones for RPs. This approach allows the model to learn the vagaries of RSSI visibility across different smartphones. In the next section, we present our experimental results, including the performance of the trained model on new smartphones not used in the RSSI training data pool.

## VI. EXPERIMENTAL RESULTS

### A. Indoor paths and smartphones

The VITAL framework was evaluated in four different buildings. The RSSI fingerprints are captured along paths in each of the buildings, as shown in figure 4. The samples were collected using a granularity of 1 meter between each RP. The blue dots in figure 4 represent the RPs for fingerprint collection. The path lengths varied from 62 meters to 88 meters. Each building also had a different number of Wi-Fi access points (WAPs). Note that only a subset of the WAPs present in a building were visible at each of the RPs in that building. Each of the buildings also had a very different material composition (wood, metal, concrete) and was populated with diverse equipment, furniture, and layouts, creating different environments in which to evaluate our framework as well as other heterogeneity-resilient localization frameworks from prior work.

For our experiments, we made use of six distinct smartphones from different manufacturers, to capture RSSI data across the RPs in the four buildings and train the indoor localization frameworks. Table I summarizes the details of these smartphones. The devices operated under the latest OS available at the time of this experiment. We refer to this set of devices as 'Base' devices. The RSSI data captured across the building consisted of five RSSI data samples captured at an RP for each of the six smartphones, across the four buildings. These five values were reduced down to three by capturing their min, max, and mean values, and used to create the three channels for each element (pixel) in the fingerprint vector that is input to our VITAL framework. The RSSI data captured was split into two datasets: training (approximately 80%) and testing (approximately 20%), with results reported for the testing dataset.

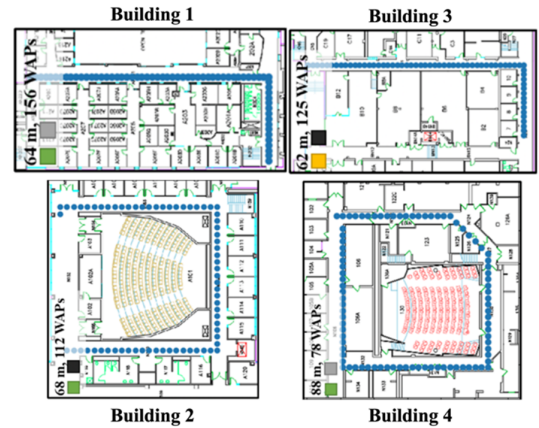

Figure 4: Benchmark indoor paths that were used for fingerprint collection. We collected RSSI data across four buildings, with varying path lengths. The blue dots indicate reference points (RP) at which the RSSI reading were taken. Six different smartphones were used to capture RSSI data at each RP.

TABLE I: SMARTPHONES USED FOR EVALUATION (BASE DEVICES).

| Manufacturer | Model | Acronym | Release Date |
|---|---|---|---|
| BLU | Vivo 8 | BLU | 2017 |
| HTC | U11 | HTC | 2017 |
| Samsung | Galaxy S7 | S7 | 2016 |
| LG | V20 | LG | 2016 |
| Motorola | Z2 | MOTO | 2017 |
| Oneplus | OnePlus 3 | OP3 | 2016 |

### B. Hyperparameter exploration

The VITAL framework involves multiple hyperparameters. While an exhaustive search across all possible values for all hyperparameters was not practically possible, we focused on analyzing four critical hyperparameters across the pre-processing and transformer architecture design, as part of a sensitivity analysis. In the first analysis, we focused on the input pre-processing phase in VITAL, within DAM. Figure 5 depicts the impact of different RSSI fingerprint image sizes and patch sizes on the mean indoor localization error with the VITAL framework.

From the results, we conclude that an RSSI fingerprint image size of R×R = 206×206 and patch size of P×P = 20×20 results in the least mean localization error. Smaller patch sizes under-perform as they lead to a greater number of patches being created, which tends to over-fit the data, whereas larger patch sizes tend to under-fit the data. The RSSI fingerprint image sizes had relatively less impact on the localization error. However, we did note that image sizes that led to the creation of partial patches at the boundaries, led to the discarding of some features, which tended to reduce accuracy.

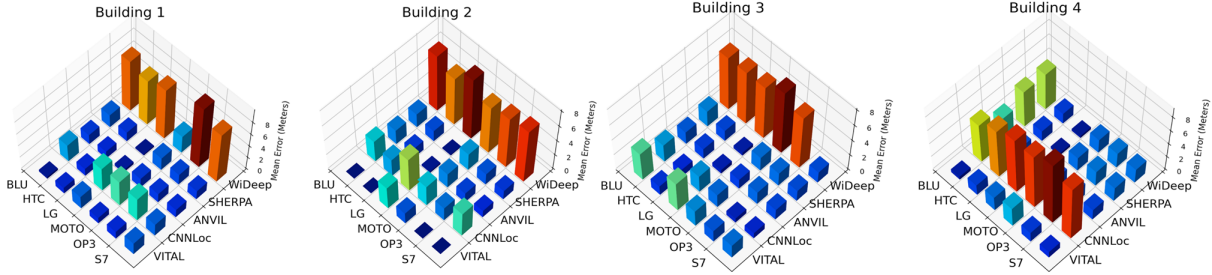

Figure 7: Mean indoor localization error across all six smartphones, four building, and localization frameworks.

We further explored two hyperparameters of the vision transformer encoder block used within the VITAL framework. We analyzed the impact of the number of multi-headed self-attention (MSA) heads and dense layers in the fine-tuning MLP block on the mean indoor localization error. Figure 6 shows a heatmap of this exploration. We observe that an MSA head count of five and two fine-tuning MLP layers show the least mean indoor localization error (shown in meters in Figure 6). A lower number of MLP layers tends to under-fit the data whereas a higher number of MLP layers tends to over-fit the data. We also observe that the number of MSA heads has a significant impact on the model's performance. A higher head count tends to over-fit the data.

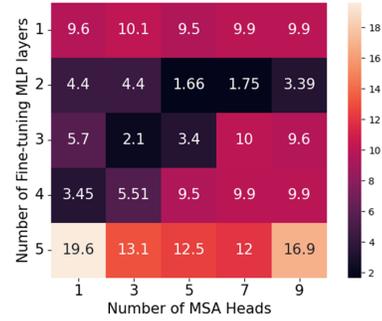

Figure 6: Impact of number of fine tuning MLP layers and number of MSA heads in our vision transformer model on mean indoor localization error.

### C. Comparison with state-of-the-art

We compared our VITAL framework with four state-of-the-art frameworks for heterogeneity-resilient indoor localization: ANVIL [19], SHERPA [20], CNNLoc [21], and WiDeep [22]. These frameworks have shown the most promising results for device heterogeneity tolerance during indoor localization. ANVIL [19] employs a multi-headed attention mechanism and a Euclidean distance-based matching approach to achieve device heterogeneity tolerance. In our proposed VITAL framework, we enhance the attention mechanism used in ANVIL using multi-headed self-attention as part of the vision transformer encoder, as explained earlier. CNNLoc [21] employs a combination of stacked autoencoder (SAE) and 1D CNN to tackle device heterogeneity. The SAE is used as an augmentation to better deal with noisy fingerprint data. In VITAL, we make use of the more powerful DAM, described earlier, to deal with noisy data. Lastly, SHERPA [20] employs a KNN algorithm enhanced with DNNs and WiDeep [22] utilizes a Gaussian Process Classifier enhanced with an SAE, respectively.

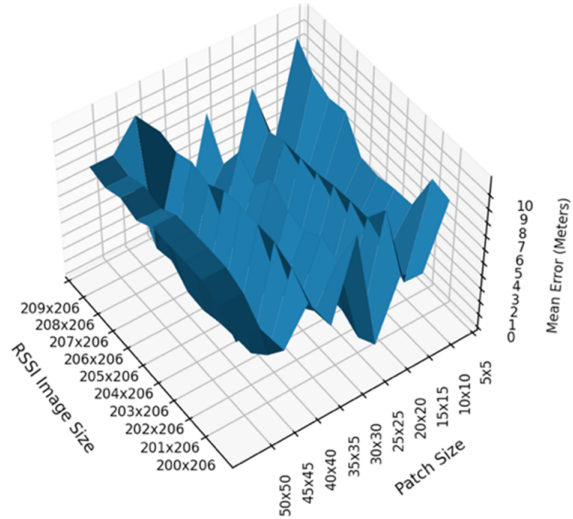

Figure 5: Surface plot showing the impact of patch size and RSSI image sizes on the mean indoor localization error.

Based on the results from our hyperparameter analysis, we finalized the configuration of the VITAL framework that was used to compare with frameworks from prior work (next Section). We used an RSSI fingerprint image size of $R×R = 206×206$, patch size of $P×P = 20×20$, number of transformer encoder blocks $L=1$, number of MSA heads = 5, number of dense layers in the MLP sub-block within the encoder = 2 (with 128 and 64 units, respectively), and the number of dense layers in the fine-tuning MLP block = 2 (with 128 and *num_classes* units, respectively). The total number of trainable parameters in the transformer model was 234,706. This compact model was able to perform inference within ~50ms, making it amenable to deployment on memory-constrained and computationally limited embedded and IoT platforms.

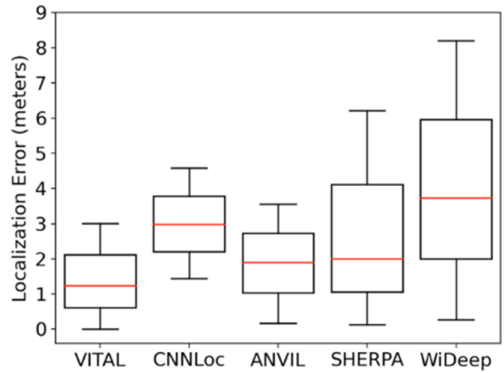

Figure 8: Min (lower whisker), Mean (red bar), and Max (upper whisker) error across all buildings for comparison frameworks w/ base devices

Figure 7 shows a color-coded comparison of VITAL with these state-of-the-art frameworks. We can observe that WiDeep consistently shows high mean errors across most buildings, making it the worst performing framework. One possible explanation for this is the effect of the de-noising SAE, which very aggressively creates noisy reconstructions of the inputs in a manner that makes it difficult for the classifier to arrive at accurate predictions. CNNLoc also employs SAE but shows better results compared to WiDeep. as its algorithms can deal with noisy data more effectively. Interestingly, CNNLoc fails to localize accurately in less noisy environments, such as that found in Building 4. ANVIL and SHERPA show better results than WiDeep and CNNLoc, with a few outliers. Our proposed VITAL framework shows overall much better results than these four state-of-the-art frameworks as it uses a more powerful learning engine (transformer) and is better able to exploit heterogeneous fingerprints captured across smartphones than the other frameworks. To better visualize the performance of these frameworks, Figure 8 shows a box plot with minimum, maximum, and mean errors across all buildings. Our VITAL framework has the least minimum, maximum, and mean errors compared to all other frameworks. VITAL has the least mean error (1.18 meters), followed by ANVIL (1.9 meters), SHERPA (2.0 meters), CNNLoc (2.98 meters), and WiDeep (3.73 meters). Thus, VITAL achieves improvements ranging from 41% to 68%. VITAL also has the lowest maximum error (3.0 meters), followed by ANVIL (3.56 meters), SHERPA (6.22 meters), CNNLoc (4.58 meters), and WiDeep (8.2 meters).

### D. Effectiveness of DAM

The Data Augmentation Module (DAM) is a significant component of the proposed VITAL framework, with broad applicability to various indoor localization frameworks. To analyze the impact of DAM, we integrated it into the four comparison frameworks. Figure 9 shows a slope graph showing the mean error for all frameworks with and without DAM. The integration of DAM allows VITAL to achieve lower mean localization errors than without it. Moreover, DAM shows notable improvements when integrated with ANVIL, SHERPA, and CNNLoc, three of the four state-of-the-art frameworks that we compare against. WiDeep shows higher mean errors with the inclusion of DAM, as it tends to overfit easily. Thus, DAM can significantly improve localization accuracy predictions for some frameworks, such as ANVIL, SHERPA, and CNNLoc, as well as our VITAL framework proposed in this paper.

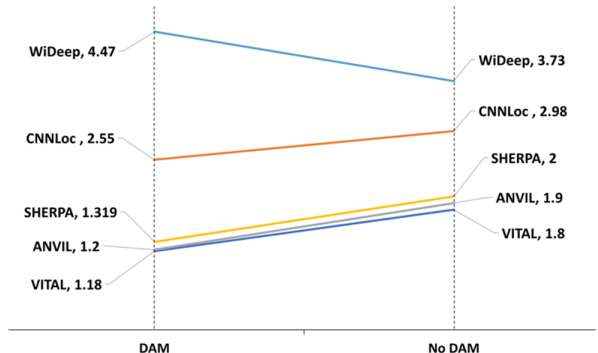
Figure 9: Slope graph of the impact of DAM on all frameworks.

### E. Results on extended (new) smartphones

To further evaluate the generalizability of our VITAL framework, we conducted another experiment by deploying the frameworks on three new smartphone devices on which the frameworks were not trained. The new devices, referred to as 'extended devices', are shown in Table II.

TABLE II: SMARTPHONES USED FOR EVALUATION (EXTENDED DEVICES).

| Manufacturer | Model | Acronym | Release Date |
|---|---|---|---|
| Nokia | Nokia 7.1 | NOKIA | 2018 |
| Google | Pixel 4a | PIXEL | 2020 |
| Apple | iPhone 12 | IPHONE | 2021 |

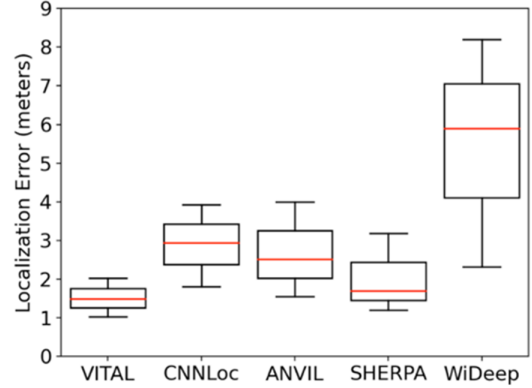
Figure 10: Min (lower whisker), Mean (red bar), and Max (upper whisker) error across all buildings for comparison frameworks w/ extended devices

Figure 10 shows a box plot with minimum, maximum, and mean errors across all buildings for the comparison frameworks, across the three extended devices. From the figure, we can conclude that VITAL outperforms all frameworks significantly. VITAL has the least mean error (1.38 meters), followed by SHERPA (1.7 meters), ANVIL (2.51 meters), CNNLoc (2.94 meters), and WiDeep (5.90 meters). Thus, VITAL achieves improvements ranging from 19% to 77%. VITAL also has the least maximum error (3.03 meters), followed by SHERPA (3.18 meters), ANVIL (4.0 meters), CNNLoc (3.92 meters), and WiDeep (8.20 meters). Thus, VITAL enables superior device heterogeneity tolerance and localization accuracy, even on new and unseen devices that it has not been trained on.

## VII. CONCLUSION

In this paper, we presented a novel framework called VITAL that is resilient to device heterogeneity during indoor localization. VITAL was compared with four state-of-the-art frameworks for device heterogeneity-resilient indoor localization, across four different buildings, and with nine different smartphone devices. VITAL shows 41% to 68% mean localization error improvement compared to the best-performing frameworks from prior work. VITAL also shows great generalizability across new smartphone devices, with 19% to 77% mean localization error improvement compared to the best-performing frameworks from prior work. The calibration-free approach enabled by VITAL makes it a preferred framework for indoor localization in real-world settings.